\documentclass[letterpaper]{article} 

\ifdefined\aaaianonymous
\usepackage[submission]{aaai2026} 
\else
\usepackage{aaai2026} 
\fi

\usepackage{times} 
\usepackage{helvet} 
\usepackage{courier} 
\usepackage[hyphens]{url} 
\usepackage{graphicx} 
\usepackage{natbib} 
\usepackage{caption} 
\usepackage{algorithm}
\usepackage{algorithmic}

\usepackage{newfloat}
\usepackage{listings}
\DeclareCaptionStyle{ruled}{labelfont=normalfont,labelsep=colon,strut=off} 
\floatstyle{ruled}
\newfloat{listing}{tb}{lst}{}
\floatname{listing}{Listing}

\usepackage{booktabs} 
\usepackage{amsmath}
\usepackage{makecell}
\usepackage{multirow}
\usepackage{times}
\usepackage{helvet}
\usepackage{courier}
\usepackage{xcolor}
\usepackage{tcolorbox}

\title{From Detection to Diagnosis: Advancing Hallucination Analysis with Automated Data Synthesis}

\author{
Yanyi Liu\textsuperscript{\rm 1}, Qingwen Yang\textsuperscript{\rm 1}, Tiezheng Guo\textsuperscript{\rm 1}, Feiyu Qu\textsuperscript{\rm 2}, Jun Liu\textsuperscript{\rm 1}, Yingyou Wen\textsuperscript{\rm 1}
\thanks{Corresponding author}
}
\affiliations
{
\textsuperscript{\rm 1}School of Computer Science and Engineering, Northeastern University, Shenyang 110819, China\\ \textsuperscript{\rm 2}Department of Computer Science, Dartmouth College, Hanover, NH 03755, USA\\ \{2290175, 2290182, 2310725\}@stu.neu.edu.cn, feiyu.qu.gr@dartmouth.edu, liujun@cse.neu.edu.cn, wenyingyou@mail.neu.edu.cn }

\begin{document}
\maketitle

\begin{abstract}
    Hallucinations in Large Language Models (LLMs), defined as the generation of content inconsistent with facts or context, represent a core obstacle to their reliable deployment in critical domains. Current research primarily focuses on binary "detection" approaches that, while capable of identifying hallucinations, fail to provide interpretable and actionable feedback for model improvement, thus limiting practical utility. To address this limitation, a new research paradigm is proposed, shifting from "detection" to "diagnosis". The Hallucination Diagnosis Task is introduced, a task which requires models to not only detect hallucinations, but also perform error localization, causal explanation, and content correction.
    We develop the Hallucination Diagnosis Generator (HDG), an automated pipeline that systematically generates high-quality training samples with rich diagnostic metadata from raw corpora through multi-dimensional augmentation strategies including controlled fact fabrication and reasoning chain perturbation. Using HDG-generated data, we train HDM-4B-RL, a 4-billion-parameter hallucination diagnosis model, employing Group Relative Policy Optimization (GRPO) with a comprehensive reward function incorporating structural, accuracy, and localization signals.
    Experimental results demonstrate that our model surpasses previous state-of-the-art detection models on the HaluEval benchmark while achieving comparable performance to advanced general-purpose models. In comprehensive diagnosis tasks, HDM-4B-RL matches the capabilities of larger general models while maintaining a smaller size. This work validates the feasibility and value of hallucination diagnosis, providing an effective methodology for building more trustworthy and reliable generative AI systems.
\end{abstract}

\ifdefined\aaaianonymous
\else
\fi

\section{Introduction}

Large Language Models (LLMs) frequently generate content that appears plausible but is factually inconsistent with world knowledge or the provided source context, this phenomenon is known as "hallucination."~\cite{ji2023survey}

This phenomenon critically undermines the reliable application of generative AI systems~\cite{chen2024systems}, particularly in high-risk domains like medical diagnosis~\cite{kim2025medical} or legal judgment~\cite{HERRERATAPIAS20251184} where outputs can inform critical decisions.
Consequently, developing robust methods to identify and mitigate hallucinations is imperative for the trustworthy deployment of LLMs.

This work focuses on "faithfulness hallucination,"~\cite{10.1162/coli.a.16} a critical challenge where a model's output contradicts or lacks support from the provided source documents in tasks that demand strict fidelity, such as long-document summarization or knowledge-based QA.

A common approach is post-hoc detection~\cite{liu2025reducing}. These methods often frame detection as a Natural Language Inference (NLI) task, assigning a coarse-grained label (e.g., Entailment, Contradiction) to the relationship between the response and its source.

\begin{figure}[tb]
    \centering
    \includegraphics[width=0.8\columnwidth]{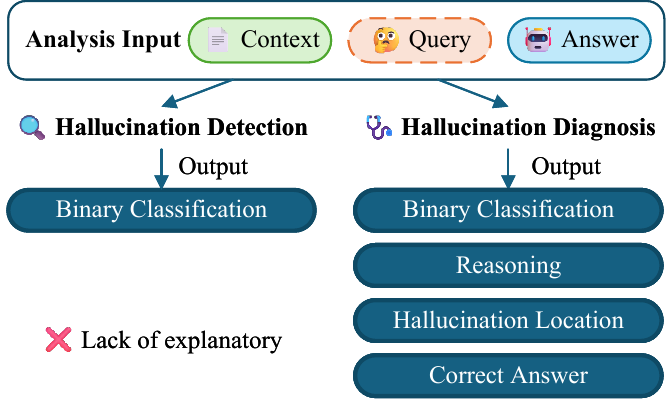}
    \caption{
        Contrasting hallucination detection with diagnosis. }
    \label{fig:diag_vs_det}
\end{figure}

However, we argue that the practical utility of this approach is limited. While a single classification label can flag an error, it fails to provide the fine-grained, interpretable feedback necessary for automated model correction or human-assisted revision. This limitation hinders the effective iterative refinement of generative models and fails to adequately address the core issue of user trust.
Given the inherent challenges of eradicating hallucinations from within a model~\cite{xu2024hallucination}, we advocate for elevating the research paradigm from mere "detection" to comprehensive "diagnosis", as shown in Figure~\ref{fig:diag_vs_det}.

Analogous to medical diagnosis, an effective diagnostic system must not only determine whether a problem exists but also offer profound insights. We therefore define the "Hallucination Diagnosis Task," which requires a model to possess core capabilities across four dimensions:

\begin{itemize}

    \item \textbf{Detection:} Accurately identifying if the generated content aligns with the source context.

    \item \textbf{Localization:} Pinpointing the specific segments within the output that are hallucinatory.

    \item \textbf{Explainability:} Explaining why the content is inconsistent.

    \item \textbf{Mitigation:} Proposing corrections or directly generating revised content that aligns with the source, based on the localization and explanation.
\end{itemize}

To facilitate this task, we began with the data, designing and implementing an automated pipeline for constructing a hallucination diagnosis dataset.
Unlike methods that rely on existing NLI or QA datasets, our pipeline samples directly from large-scale pre-training corpora and applies multi-dimensional enhancement strategies—including controlled fact forgery, reasoning chain perturbation, and ambiguous information replacement, to automatically generate diverse diagnostic samples covering various task types and difficulty levels.

Finally, using the data produced by this pipeline, a 4B-scale hallucination diagnosis model, was trained and its effectiveness validated.

The main contributions of this work are as follows:
\begin{itemize}

    \item A novel task, Hallucination Diagnosis Task, is proposed and formally defined with its core capabilities.

    \item The design and implementation of an automated pipeline to construct diagnostic datasets from large-scale pre-training corpora.

    \item A diagnosis model, HDM-4B-RL, is trained on the generated dataset, has proven effective across key diagnostic capabilities, thereby validating the proposed methods and data.
\end{itemize}

\section{Related Work}

\subsection{Hallucination Detection}

Existing methods for hallucination detection can be broadly categorized into two paradigms.

One line of research focuses on developing cost-efficient, specialized factuality classifiers. For example, methods like SummaC~\cite{laban2022summac} adapt Natural Language Inference (NLI) models for document-level consistency checking. Other approaches, such as QAFactEval~\cite{fabbri2022qafacteval}, leverage Question Answering (QA) frameworks to verify factual claims, while MiniCheck~\cite{tang2024minicheck} utilizes LLM-generated data for training its classifier, achieving state-of-the-art performance on real-world hallucination benchmarks.

The second paradigm leverages the inherent reasoning capabilities of Large Language Models (LLMs) for verification, often by decomposing the complex detection task into more manageable sub-steps. For instance, CoNLI~\cite{lei2023chain} decomposes a claim and applies NLI to each segment. RaRR~\cite{gao2022rarr} verifies factuality by generating relevant questions coupled with external retrieval. Lynx~\cite{ravi2024lynx} utilizes the model's intrinsic reasoning capabilities to construct a factuality detection model through fine-tuning.

The present work moves beyond this binary fact-checking to introduce the more comprehensive task of "Hallucination Diagnosis."
This task requires a model to not only detect an error but also to localize its position, provide natural language reasoning, and suggest a correction. We enable this integrated diagnostic capability through a novel, automated pipeline that synthesizes richly annotated diagnostic data from large-scale corpora.

\subsection{Reasoning in Large Language Models}

Significant progress has recently been made in eliciting and steering the reasoning capabilities of Large Language Models (LLMs).
The seminal Chain-of-Thought (CoT) prompting method~\cite{wei2022chain} demonstrated that prompting models to generate step-by-step thought processes dramatically improve their performance on complex tasks.

This insight has spurred a wealth of follow-up research. For example, Self-Consistency~\cite{wang2022self} enhances robustness by sampling diverse reasoning paths and taking a majority vote, while Tree-of-Thoughts (ToT)~\cite{yao2023tree} generalizes this into a tree-like exploration that enables deeper planning and backtracking.

More recent advancements have pushed beyond prompt engineering to innovate on the models' intrinsic capabilities. Models such as DeepSeek-R1~\cite{guo2025deepseek}, for instance, perform deep reasoning before generating a response, significantly enhancing their performance on complex tasks.
Collectively, these works establish that modern LLMs possess a continuously evolving, intrinsic capacity for executing complex, multi-step logical operations.

This study builds directly on this progress, aiming to apply this powerful reasoning capability to the analysis of a model's own outputs, which is the core of the hallucination diagnosis task. The maturity and ongoing evolution of these reasoning capabilities are the key foundations that make this ambitious task technically feasible.

\section{Methodology}

This section details our proposed methodology for hallucination diagnosis. The approach comprises two core components: (1) the Hallucination Diagnosis Generator (HDG), a data synthesis pipeline, and (2) the Hallucination Diagnosis Model (HDM), a model trained on the data produced by HDG.

\subsection{Hallucination Diagnosis Generator (HDG)}

\begin{figure}[tb]
    \centering
    \includegraphics[width=\columnwidth]{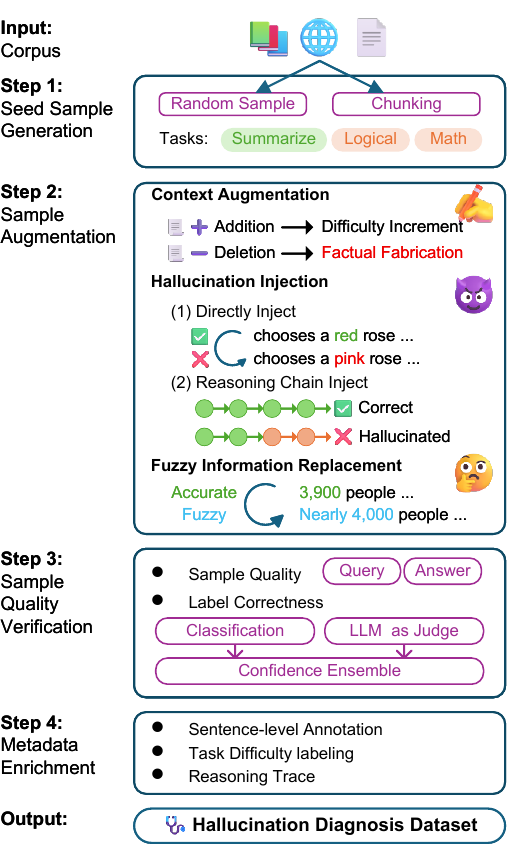}
    \caption{An overview of the Hallucination Diagnosis Generator (HDG) pipeline, detailing its four core stages from raw text to richly annotated diagnostic data.
    }
    \label{fig:main}
\end{figure}

This section provides a detailed overview of our automated pipeline for constructing a hallucination diagnosis dataset, named the Hallucination Diagnosis Generator (HDG).
The pipeline is designed to systematically generate a large-scale, high-quality, and richly annotated training dataset from raw pre-training corpora, specifically for training hallucination diagnosis models.
Unlike datasets adapted from existing QA and NLI datasets, our pipeline generates more diverse samples with a broader difficulty distribution.
As illustrated in Figure~\ref{fig:main}, the pipeline consists of four stages: task-oriented seed sample generation, multi-dimensional sample augmentation, quality verification, and metadata enrichment.

\paragraph{Step 1: Task-Oriented Seed Sample Generation}
The objective of this stage is to generate high-quality "seed samples" from a large, unstructured corpus. Each sample, comprising a context, a query, and an answer, serves as the foundation for subsequent augmentation.

First, reference texts are sampled from a pre-training corpus and filtered using heuristic strategies (e.g., removing paragraphs that are too short or have high perplexity) to ensure informational richness and linguistic quality. For long documents, a recursive text splitter is employed to generate shorter context samples.

Diverse instructions are generated for the sampled texts, covering tasks such as summarization, logical reasoning, and mathematics to ensure a wide range of task types and difficulty levels.

Finally, responses for each instruction are generated using LLMs. For logical reasoning and mathematical tasks, we leverage the model's reasoning capabilities to maximize the accuracy of the generated answers.
Specifically, we prompt the model to produce responses with a Chain-of-Thought (CoT) structure. The resulting seed samples are generated in two formats: one with a direct answer and another including the full reasoning chain.

\paragraph{Step 2: Sample Augmentation}
This stage represents the core innovation of our pipeline, designed to augment seed samples through a series of controllable perturbation strategies.
Three primary augmentation strategies are employed: context augmentation, hallucination injection, and fuzzy information replacement.

\textbf{Context Augmentation:} Samples are enhanced by programmatically adding or deleting information in the context. Information addition introduces distractors to increase the difficulty of extracting relevant facts. Conversely, information deletion removes details pertinent to the original answer, thereby creating controlled scenarios of factual inconsistency.

\textbf{Hallucination Injection:} This strategy directly introduces errors into "correct" responses via two methods: 1) \textit{Direct Injection}, applied to tasks like summarization, which creates factual errors by replacing key entities (e.g., "red rose" to "pink rose"); and 2) \textit{Reasoning Chain Injection}, applied to logical and mathematical tasks, which generates logical hallucinations by perturbing a key step in a correct reasoning chain.

\textbf{Fuzzy Information Replacement:} This approach replaces precise information (e.g., 3,900 people) with vague expressions (e.g., nearly 4,000 people). This process creates outputs that are semantically plausible but not strictly faithful to the source. This type of augmentation is applied only to summarization tasks, as such ambiguity is unacceptable in reasoning and mathematical contexts.

\paragraph{Step 3: Sample Quality Verification}
This critical quality control stage ensures that augmented data pairs are linguistically fluent, logically coherent, and accurately labeled as either containing a hallucination or not.

Quality control involves two evaluation dimensions: \textbf{Task Sample Quality}, where an LLM scores the query-answer pair for clarity, fluency, and coherence (while ignoring correctness); \textbf{Hallucination Label Correctness}, where we use an ensemble-based approach to enhance labeling accuracy by integrating judgments from multiple models.
An ensemble of top-tier classifiers (e.g., Lynx) and LLM-as-a-Judge methods~\cite{gu2024survey} with a strong model (e.g., Qwen3-32B) integrate judgments based on confidence scores to ensure accuracy.

\paragraph{Step 4: Metadata Enrichment}
After quality verification, samples are enriched with structured metadata, transforming them from simple detection examples into comprehensive data suitable for training diagnostic models.

Core diagnostic labels include: \textbf{Sentence-Level Annotation}, which marks the specific location of hallucinations to train the model's localization capability; \textbf{Task Difficulty Labels}, which assigns a difficulty level based on the reasoning chain length required for the model to answer the original instruction; and \textbf{Reasoning Trace}, which provides an explanatory chain for the final judgment.

The final output is a structured dataset where each entry contains the source context, instruction, response, hallucination location, difficulty, and a ground-truth answer, providing a robust foundation for training the HDM.

\subsection{Hallucination Diagnosis Model}

Fearing SFT could disrupt the base model's hybrid reasoning, we employed the Group Relative Policy Optimization (GRPO)~\cite{shao2024deepseekmath} algorithm, using RL to guide the model's diagnostic capabilities while preserving its core abilities.

GRPO updates its policy by comparing the relative quality of a group of generated samples rather than relying on a single absolute score. This approach enhances training efficiency and stability while helping to preserve the model's native capabilities during the alignment process.

To translate the abstract task of "hallucination diagnosis" into a concrete, learnable objective for the model, a comprehensive, rule-based reward function is designed.
This function assesses the model's output across multiple dimensions to guide it toward the desired diagnostic behaviors. The reward system consists of the following key components:

\paragraph{Structured Output Reward ($R_{struct}$)}
This foundational reward ensures a well-formed, machine-readable output, which is required to make other diagnostic components parsable. The model is incentivised to generate its diagnostic report as a JSON object containing four key fields: \textit{conclusion}, \textit{diagnosis}, \textit{hallucinations}, and \textit{corrected\_answer}.

Each field is scored independently, and the final reward is the average of these scores. An incomplete or malformed JSON structure yields a reward of 0.

\paragraph{Detection Accuracy Reward ($R_{acc}$)}
This component treats hallucination detection as a binary classification task. It evaluates the accuracy of the model's judgment in the `conclusion` field regarding whether the original answer contains hallucinations.

A positive reward is granted when the model's judgment aligns with the ground-truth label, calculated as follows:
\begin{equation}
    R_{acc}= I(y = \hat{y})
\end{equation}
where $I(\cdot)$ is the indicator function (1 if true, 0 otherwise), $y$ is the model's predicted conclusion (e.g., "pass" for no hallucination, "fail" for hallucination), and $\hat{y}$ is the corresponding ground-truth label.

\paragraph{Localization Reward ($R_{loc}$)}
This reward function evaluates the model's ability to locate and delineate hallucinatory content. The reward is calculated based on the degree of overlap between the set of predicted hallucinatory sentences and the ground-truth set.

Let $S_{pred}$ be the set of sentences predicted as hallucinatory, and $S_{gt}$ be the ground-truth set. A valid "hit" occurs when a predicted sentence $s_{p}\in S_{pred}$ has a string containment relationship with a ground-truth sentence $s_{gt}\in S_{gt}$ (i.e., $s_{p}$ is a substring of $s_{gt}$, or vice versa).

In the event of a hit, a partial score is calculated based on the length ratio of the two sentences to penalize imprecise boundaries. Specifically, the scoring function $\text{score}(s_{p}, s_{gt})$ is defined as:
\begin{equation}
    \text{score}(s_{p}, s_{gt}) =
    \begin{cases}
        \min\left(\frac{L_p}{L_{gt}}, \frac{L_{gt}}{L_p}\right) & \text{if }\text{hit}(s_{p}, s_{gt}) \\
        0                                                       & \text{otherwise}
    \end{cases}
\end{equation}
where $L_{p}$ and $L_{gt}$ are the lengths of the predicted and ground-truth sentences, respectively. The score is 0 if there is no hit.

The final localization reward $R_{loc}$ is defined as the sum of scores for each predicted sentence against its best-matching ground-truth sentence, normalized by the total number of ground-truth sentences to account for recall:
\begin{equation}
    R_{loc}= \frac{1}{|S_{gt}|}\sum_{s_p \in S_{pred}}\max_{s_{gt} \in S_{gt}}\left\{ \text{score}(s_{p}, s_{gt}) \right\}
\end{equation}

\paragraph{Final Reward Aggregation}
The final total reward $R$ is the weighted sum of the above components, defined as:
\begin{equation}
    R = \alpha_{1}R_{struct}+ \alpha_{2}R_{acc}+ \alpha_{3}R_{loc}
\end{equation}
The hyperparameters $\alpha_{1}, \alpha_{2}$ and $\alpha_{3}$ are set to 1.0, 0.5, and 0.5, respectively, to balance the contribution of each component to the overall reward.

\section{Experiments}

We designed a series of experiments to validate the effectiveness of our methodology across three core tasks: hallucination detection, localization, and mitigation.

\begin{figure*}[ht]
    \centering
    \includegraphics[width=\textwidth]{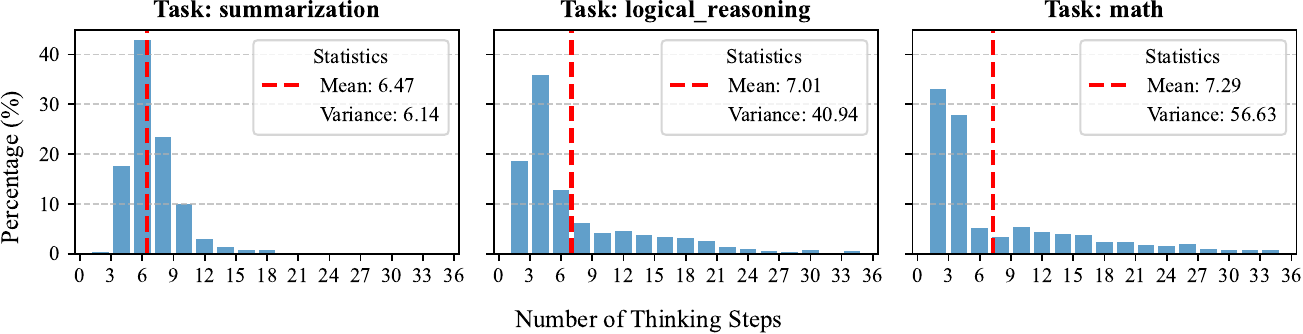}
    \caption{Distribution of reasoning steps in the synthesized hallucination diagnosis dataset, categorized by task type.}
    \label{fig:difficulty_distribution}
\end{figure*}

\subsection{Evaluation Datasets}

Our evaluation framework covers two primary task categories: Question-Answering (QA) and Summarization.
For QA tasks, we employ all subsets from the HaluBench benchmark~\cite{ravi2024lynx}, which spans topics from general knowledge to specialized domains like finance and medicine.
For the summarization task, we use the SummEval~\cite{fabbri2020summeval} for benchmarking. Detailed statistics for these datasets are presented in Table \ref{tab:datasets}.

\begin{table}[tb]
    \centering

    \begin{tabular}{lccc}
        \toprule \textbf{Dataset} & {\textbf{Total}} & {\textbf{Halu}} & {\textbf{Non-Halu}} \\
        \midrule
        \multicolumn{4}{l}{\textit{QA Task (HaluBench)}}                                     \\
        \quad HaluEval (HE)       & 10000            & 5010            & 4990                \\
        \quad RAGTruth (RT)       & 1000             & 160             & 740                 \\
        \quad FinanceBench (FB)   & 1000             & 500             & 500                 \\
        \quad DROP                & 1000             & 500             & 500                 \\
        \quad CovidQA (CQA)       & 1000             & 500             & 500                 \\
        \quad PubMedQA (PMQA)     & 1000             & 500             & 500                 \\
        \midrule
        \multicolumn{4}{l}{\textit{Summarization Task}}                                      \\
        \quad SummEval (SE)       & 1600             & 294             & 1306                \\
        \bottomrule
    \end{tabular}
    \caption{Distribution of samples in the evaluation datasets. The name in parentheses indicates the abbreviation used in subsequent sections.}
    \label{tab:datasets}
\end{table}

In addition to detection, we also evaluate the full scope of hallucination diagnosis, which includes localizing and mitigating hallucinatory content.

To this end, we constructed a dedicated diagnosis benchmark from the SummEval dataset. We first grouped all generated summaries by their reference text. From these groupings, we filtered for all summaries identified as containing hallucinations and further annotated these samples to pinpoint the specific sentences that were hallucinatory.

This process resulted in a diagnosis dataset of 285 samples, each containing a consistency label, sentence-level hallucination annotations, and ground-truth responses.

\subsection{Experimental Setup}

\subsubsection{Data Construction}

We began by using our proprietary Hallucination Diagnosis Generator (HDG) pipeline to construct the training dataset.
The data was sourced from the Wikipedia (20231101.en) dump, chosen for its breadth in establishing a general-purpose methodology, from which we randomly sampled 4,500 documents.
Within the HDG pipeline, different models were employed at various stages based on a cost/efficiency/quality trade-off (e.g., efficient models for generation, strong ensembles for annotation).

The task generation stage utilizes Gemini-2.5-flash-lite, while the corresponding responses and injections are generated by Qwen3-32B~\cite{yang2025qwen3}.
The context augmentation stage employs Qwen3-Embedding-0.6B~\cite{zhang2025qwen3} as an embedding model for similarity retrieval.
In the quality verification stage, the judgment ensemble included Qwen3-32B~\cite{yang2025qwen3}, GPT-4.1, Lynx~\cite{ravi2024lynx}, and MiniCheck-7B~\cite{tang2024minicheck}. The subsequent metadata enrichment was performed by Qwen3-32B.

\begin{figure}[tb]
    \centering
    \includegraphics[width=0.95\columnwidth]{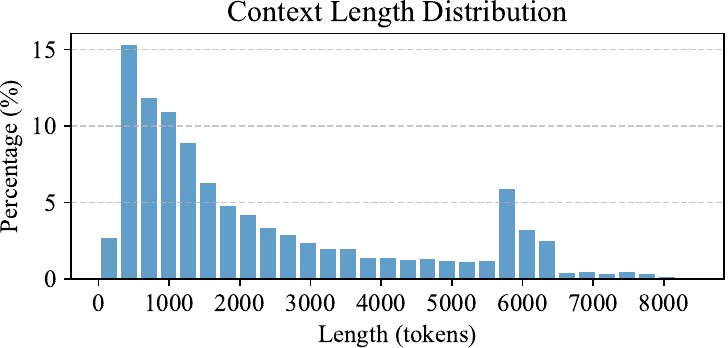}
    \caption{Distribution of context lengths in the constructed hallucination diagnosis dataset.}
    \label{fig:context_length}
\end{figure}

\begin{table}[tb]
    \centering

    \begin{tabular}{lrrr}
        \toprule
        \textbf{Task}           & \textbf{Raw}  & \textbf{Avg}   & \textbf{Total}                \\
        \midrule Summary        & 2332          & 5308           & 7640 (3236 / 4404)            \\
        Logical                 & 988           & 6205           & 7193 (1613 / 5580)            \\
        Math                    & 719           & 2901           & 3620 (1150 / 2470)            \\
        \midrule \textbf{Total} & \textbf{4039} & \textbf{14414} & \textbf{18453 (5999 / 12454)} \\
        \bottomrule
    \end{tabular}
    \caption{Statistics of the constructed hallucination diagnosis dataset. The numbers in parentheses indicate the number of hallucination and non-hallucination samples, respectively.}
    \label{tab:task_stats_standard}
\end{table}
This process yielded a dataset of 18,453 samples, with the distribution across task types shown in Table \ref{tab:task_stats_standard}.
Analyses of the dataset's context length and difficulty distribution are presented in Figures \ref{fig:context_length} and \ref{fig:difficulty_distribution}, respectively.

The difficulty, measured by reasoning steps, increases from summarization to logical and mathematical reasoning. The latter two tasks also exhibit a larger variance, indicating a broader range of complexity.

\subsubsection{Model Training}

Qwen3-4B is the base model for training HDM-4B-RL.
The training was conducted using the ms-swift~\cite{zhao2025swift} and Hugging Face Transformers~\cite{wolf-etal-2020-transformers} libraries, employing the Group Relative Policy Optimization (GRPO) algorithm.
We utilized 2 $\times$ NVIDIA A100 80GB GPUs. During training, one GPU was dedicated to the policy model for rollouts, while the other handled reward computations and policy model parameter updates.
For training hyperparameters, The AdamW optimizer is used with a learning rate of $1 \times 10^{-6}$ and a weight decay of 0.01. The he model was trained for a total of 2306 steps in batch-size 64.

\begin{table*}
    [htbp]
    \centering

    \begin{tabular}{lccccccccc}
        \toprule \textbf{Model}           & \textbf{Size} & \textbf{HE}       & \textbf{RT}       & \textbf{FB}       & \textbf{DROP}     & \textbf{CQA}      & \textbf{PMQA}     & \textbf{SE}       & \textbf{Average}  \\
        \midrule \multicolumn{10}{l}{\textit{LLM Prompt Based}}                                                                                                                                                           \\
        \midrule Qwen3-4B (non-reasoning) & 4B            & 73.64             & 58.25             & 56.72             & 52.29             & 78.26             & 68.83             & 66.43             & 64.92             \\
        Qwen3-4B                          & 4B            & 75.15             & 61.28             & 82.98             & 76.19             & 87.61             & 77.72             & 76.26             & 76.74             \\
        Qwen3-32B (non-reasoning)         & 32B           & 80.92             & 57.83             & 69.59             & 70.61             & 83.94             & 83.43             & 72.61             & 74.13             \\
        Qwen3-32B                         & 32B           & 77.63             & \underline{69.36} & \underline{86.58} & \textbf{80.10}    & \underline{92.99} & \underline{87.10} & \textbf{78.16}    & \underline{81.70} \\
        GPT-4.1                           & -             & \underline{82.95} & 62.58             & 60.79             & 66.91             & 90.75             & \textbf{88.17}    & 75.41             & 75.37             \\
        o4-mini                           & -             & \textbf{83.23}    & \textbf{71.89}    & \textbf{88.90}    & \underline{76.28} & \textbf{94.19}    & 85.95             & \underline{77.88} & \textbf{82.62}    \\
        \midrule \multicolumn{10}{l}{\textit{Hallucination Detection Models}}                                                                                                                                             \\
        \midrule HHEM                     & 110M          & 66.59             & \textbf{73.72}    & 40.85             & 46.55             & 57.48             & 57.22             & 61.00             & 57.63             \\
        Alignscore-Large                  & 355M          & 74.10             & 56.77             & 41.93             & 50.90             & 64.11             & 61.05             & 65.60             & 59.21             \\
        FactCG-DeBERTa-v3-Large           & 435M          & 66.29             & 52.28             & 42.03             & 48.67             & 78.48             & 59.56             & 61.00             & 58.33             \\
        MiniCheck-Flan-T5-Large           & 783M          & 66.85             & 53.65             & 42.46             & 49.74             & 75.48             & 71.72             & \underline{76.20} & 62.30             \\
        Bespoke-MiniCheck-7B              & 7B            & 79.73             & 53.08             & 50.38             & 64.78             & 88.10             & 64.75             & \textbf{78.56}    & 68.48             \\
        Lynx-8B                           & 8B            & \textbf{86.96}    & 60.25             & \underline{74.30} & \underline{64.79} & \textbf{97.70}    & \textbf{88.37}    & 68.39             & \underline{77.25} \\
        HDM-4B-RL (non-reasoning)         & 4B            & 82.84             & 63.74             & 57.83             & 50.85             & 84.21             & 80.67             & 70.75             & 70.13             \\
        HDM-4B-RL                         & 4B            & \underline{83.96} & \underline{68.52} & \textbf{84.54}    & \textbf{74.39}    & \underline{92.50} & \underline{77.79} & 75.88             & \textbf{79.65}    \\
        \bottomrule
    \end{tabular}

    \caption{Hallucination Detection Results. Macro F1 score is used for comparison. Bold and underlined values indicate the best and the second-best results in each group, respectively.}

    \label{tab:hallucination_detection}
\end{table*}

\subsubsection{Evaluation Setup}

For evaluation, open-source LLMs were deployed on a single NVIDIA A100 80GB GPU using the vLLM framework~\cite{kwon2023efficient}.
For models too large for a single GPU, such as Qwen3-32B, we applied FP8 quantization to enable deployment.
Proprietary models were evaluated via their official API endpoints.
Classifier-based models were loaded using the Transformers library and evaluated with their default inference configurations.

\subsection{Hallucination Detection}

To validate the capabilities of our model HDM-4B-RL, we conducted evaluations on the hallucination detection task. Following the HaluBench benchmark, we treat the hallucination detection task as a binary classification task, and use the macro F1 score as the primary metric.

For a comprehensive comparison, our evaluation includes two categories of baselines: (1) specialized hallucination detection models, including HHEM~\cite{hhem-2.1-open}, Alignscore~\cite{zha-etal-2023-alignscore}, FactCG~\cite{lei2025factcg}, MiniCheck~\cite{tang2024minicheck}, and Lynx~\cite{ravi2024lynx}; and (2) prompt-based LLMs, for which we use Qwen3-4B, Qwen3-32B, GPT-4.1, and o4-mini.
For models with a switchable reasoning mode, like Qwen3, we report results for both modes.
The results are presented in Table \ref{tab:hallucination_detection}.

\subsubsection{Comparison with Hallucination Detection Models}
The results show that HDM-4B-RL is the top-performing specialized hallucination detection model, excelling on FinanceBench and DROP with F1 scores of 84.54 and 74.39, respectively. Compared to the previous SOTA model, Lynx-8B, our model achieved a 2.4\% higher average F1 score despite being half the size. This demonstrates our method's superior balance between computational efficiency and performance, achieving better results with a smaller model.

\subsubsection{Comparison with General LLMs}
We also acknowledge that a performance gap remains when compared to top-tier prompt-based models like o4-mini (82.62 F1).
However, our approach shows immense potential. The performance of HDM-4B-RL (79.65 F1) already surpasses powerful closed-source models like GPT-4.1 (75.37 F1) and is closing the gap with the reasoning-enabled Qwen3-32B (81.70 F1). This indicates that through specialized fine-tuning, a smaller model can challenge unoptimized larger models on specific tasks, offering an efficient solution for resource-constrained scenarios. This is highlighted in detection latency, where the 32B model requires approximately 2.2$\times$ the runtime of ours.

\subsubsection{Reasoning Mode vs. Non-Reasoning Mode}
The performance benefit of the reasoning mode is a key finding of our experiments. For the Qwen3-4B model, enabling its reasoning mode boosted the average F1 score from 64.92 to 76.74.
This effect was most pronounced on datasets like FinanceBench (heavy on calculation) and DROP (reliant on discrete reasoning), which align with the strengths of reasoning-focused models.
This enhancement effect is equally significant on our fine-tuned model, HDM-4B-RL, with the F1 score rising from 70.13 (non-reasoning) to 79.65 (reasoning). Given that we performed RL fine-tuning only in reasoning mode, this indicates that our method does not compromise the model's performance in non-reasoning mode.
Reasoning mode requires approximately $1.8\times$ the runtime of non-reasoning mode, presenting an accuracy-efficiency trade-off adaptable to specific needs.

\subsubsection{Effectiveness of RL Training}

Table \ref{tab:hdm_comparison_basic} provides direct evidence of the effectiveness of our fine-tuning. Compared to its base model, Qwen3-4B, HDM-4B-RL achieves comprehensive improvements.
A deeper analysis reveals the source of this performance boost. In reasoning mode, the F1 score improvement (+2.91) is primarily driven by a significant gain in Recall (+4.62) and a solid increase in Precision (+2.50). This indicates that our model's ability to "identify all true hallucinations" has been greatly enhanced without sacrificing judgmental accuracy.
More interestingly, in non-reasoning mode, the improvement in the F1 score is even larger (+5.21). This is driven almost entirely by a massive leap in Recall (+5.97), while Precision remains stable (-0.26). This suggests our fine-tuning strategy greatly enhanced the model's fundamental pattern recognition capabilities, making it more sensitive to capturing the features of hallucinations even without relying on complex reasoning.

\begin{table}[tb]
    \centering

    \begin{tabular}{lcccc}
        \toprule \textbf{Model}    & \textbf{P}      & \textbf{R} & \textbf{F1} & \textbf{Acc} \\
        \midrule Lynx-8B           & 77.79           & 77.54      & 77.25       & 81.63        \\
        \midrule Qwen3-4B$^{\dag}$ & 70.93           & 65.35      & 64.92       & 71.83        \\
        HDM-4B-RL$^{\dag}$         & \makecell{70.67                                           \\ (-0.26)} & \makecell{71.32 \\ (+5.97)} & \makecell{70.13 \\ (+5.21)} & \makecell{73.74 \\ (+1.91)} \\
        \midrule Qwen3-4B          & 77.89           & 76.42      & 76.74       & 80.85        \\
        HDM-4B-RL                  & \makecell{80.39                                           \\ (+2.50)} & \makecell{81.04 \\ (+4.62)} & \makecell{79.65 \\ (+2.91)} & \makecell{82.24 \\ (+1.39)} \\
        \bottomrule
    \end{tabular}
    \caption{Performance comparison with Qwen3-4B. P, R, F1, and Acc represent Precision, Recall, F1 score, and Accuracy, respectively. The $^{\dag}$ indicates evaluation in non-reasoning mode.}
    \label{tab:hdm_comparison_basic}
\end{table}

\subsection{Hallucination Diagnosis Result}

Lacking established public benchmarks for the full diagnosis task, the hallucination diagnosis task is conceptualized as a composite challenge, and its evaluation is conducted by deconstructing it into three core sub-tasks: detection, localization, and mitigation.

For the granular analysis, specific metrics are applied to each sub-task. For \textbf{detection}, given that this evaluation exclusively involves samples containing hallucinations, performance is measured directly by Accuracy~(ACC). The \textbf{localization} capability is assessed by a "Hit Rate"~(HR), analogous to the $R_{loc}$ reward, and supplemented by a "Span Validity"~(SV) metric to ensure localized spans are verbatim substrings of the original answer. Finally, the quality of the corrected text from the \textbf{mitigation} sub-task is evaluated using AlignScore~\cite{zha-etal-2023-alignscore} (AS), a reference-based metric for factual consistency.

To establish performance benchmarks, two baseline approaches based on prompt engineering were constructed. The first is the single-prompt method, which attempts to generate the entire diagnostic report in a single pass, our HDM-4B-RL is a representative of this method. The second is the pipeline method, which decomposes the task into a sequence of three separate steps: detection, localization, and mitigation.
The final results are presented in Table \ref{tab:hallucination_diagnosis}

\begin{table}[tb]
    \centering

    \begin{tabular}{lcccc}
        \toprule
        \multirow{2}{*}{\textbf{Model}}              & \multicolumn{1}{c}{\textbf{Det}} & \multicolumn{2}{c}{\textbf{Loc}} & \multicolumn{1}{c}{\textbf{Mit}}                                   \\
        \cmidrule(lr){2-2} \cmidrule(lr){3-4} \cmidrule(lr){5-5}
                                                     & \multicolumn{1}{c}{\textbf{Acc}} & \multicolumn{1}{c}{\textbf{HR}}  & \multicolumn{1}{c}{\textbf{SV}}  & \multicolumn{1}{c}{\textbf{AS}} \\
        \midrule \multicolumn{4}{l}{Original Result} & 59.77                                                                                                                                    \\
        \midrule \multicolumn{5 }{l}{\textit{Pipeline Method}}                                                                                                                                  \\
        Qwen3-32B                                    & 94.74                            & 76.97                            & 69.69                            & 79.55                           \\
        GPT 4.1                                      & 61.75                            & 60.52                            & 53.08                            & 76.77                           \\
        o4-mini                                      & 82.46                            & 78.59                            & 79.71                            & 78.51                           \\
        \midrule \multicolumn{5}{l}{\textit{Single Prompt Method}}                                                                                                                              \\
        Qwen3-32B                                    & 97.54                            & 64.85                            & 59.15                            & 70.98                           \\
        GPT 4.1                                      & 77.54                            & 59.12                            & 48.10                            & 65.97                           \\
        o4-mini                                      & 89.12                            & 41.30                            & 43.04                            & 61.18                           \\
        HDM-4B-RL                                    & 92.28                            & 58.65                            & 48.49                            & 69.16                           \\
        \bottomrule
    \end{tabular}
    \caption{Performance Comparison on the Hallucination Diagnosis Task.
        Det, Loc, and Mit represent detection, localization, and mitigation tasks. The "Original Result" serves as a baseline for the mitigation task. }
    \label{tab:hallucination_diagnosis}
\end{table}

Table \ref{tab:hallucination_diagnosis} reveals a distinct trade-off. While the multi-step pipeline achieves superior performance on fine-grained tasks like localization and mitigation via task decomposition, it incurs high latency due to multiple LLM calls. Conversely, the end-to-end method is far more efficient: using HDM e2e as the baseline (1$\times$ runtime), 32B e2e requires only 1.5$\times$, whereas the 32B pipeline demands 4.9$\times$.

Against this backdrop, the proposed HDM-4B-RL model is particularly noteworthy. Operating as a single-prompt model, it demonstrates a clear and comprehensive performance advantage over other baselines in its category, including GPT-4.1 and o4-mini. More importantly, as a lightweight 4B model, its performance is highly competitive across all diagnostic sub-tasks, closely approaching the quality achieved by the 32B-scale model, even in the most challenging mitigation task. This outcome strongly validates that a lightweight, end-to-end model, when subjected to specialized alignment with high-quality data, can achieve performance comparable to that of much larger counterparts that rely on more cumbersome processes. This provides a solid basis for developing AI systems that are both highly efficient and reliable.

\section{Limitations and Conclusion}

\paragraph{Limitations}

Despite promising results, our study has several limitations that highlight avenues for future research.
First, regarding the data, our HDG pipeline currently relies solely on Wikipedia as its source corpus. Future work should expand these data sources to more specialized domains (e.g., scientific literature, legal documents) and further scale the dataset's size.
Second, due to computational constraints, our experiments were confined to a 4B-parameter model. Exploring the effectiveness of our approach on larger-scale models is a key direction for future work.

\paragraph{Conclusion}
In this paper, we addressed the problem of hallucination in large language models by proposing a "hallucination diagnosis" paradigm that moves beyond traditional binary detection.
We defined this task by four core capabilities—detection, localization, explanation, and mitigation. To tackle this, we introduced a novel methodology featuring the HDG automated data pipeline and trained an efficient Hallucination Diagnosis Model, HDM-4B-RL, using the GRPO algorithm on our synthesized data.
Experiments demonstrated that this specialized 4B model not only outperforms larger, detection-specific models but also achieves highly competitive performance in the full diagnostic workflow, rivaling that of powerful general-purpose models. These results validate the efficacy of our approach, presenting a promising path toward developing more reliable and trustworthy AI systems.

\section{Acknowledgments}
This study is supported by Liaoning Provincial Science and Technology Innovation Project in the Field of Artificial Intelligence (Project name: Research on key technologies for systems engineering of large language model)(Grant no.2023JH26/10100005).

\bibliography{aaai2026}

@article{chen2024systems,
  title     = {Systems engineering issues for industry applications of large language model},
  author    = {Chen, Wang and Yan-yi, Liu and Tie-zheng, Guo and Da-peng, Li and Tao, He and Zhi, Li and Qing-wen, Yang and Hui-han, Wang and Ying-you, Wen},
  journal   = {Applied Soft Computing},
  volume    = {151},
  pages     = {111165},
  year      = {2024},
  publisher = {Elsevier}
}

@article{ji2023survey,
  title     = {Survey of hallucination in natural language generation},
  author    = {Ji, Ziwei and Lee, Nayeon and Frieske, Rita and Yu, Tiezheng and Su, Dan and Xu, Yan and Ishii, Etsuko and Bang, Ye Jin and Madotto, Andrea and Fung, Pascale},
  journal   = {ACM Computing Surveys},
  volume    = {55},
  number    = {12},
  pages     = {1--38},
  year      = {2023},
  publisher = {ACM New York, NY}
}

@inproceedings{lei2023chain,
  title     = {Chain of natural language inference for reducing large language model hallucinations},
  author    = {Lei, Deren and Li, Yaxi and Hu, Mengya and Wang, Mingyu and Yun, Xi},
  booktitle = {NeurIPS 2023 Workshop on Instruction Tuning and Instruction Following},
  year      = {2023}
}

@inproceedings{zha-etal-2023-alignscore,
  title     = {{A}lign{S}core: Evaluating Factual Consistency with A Unified Alignment Function},
  author    = {Zha, Yuheng  and
               Yang, Yichi  and
               Li, Ruichen  and
               Hu, Zhiting},
  editor    = {Rogers, Anna  and
               Boyd-Graber, Jordan  and
               Okazaki, Naoaki},
  booktitle = {Proceedings of the 61st Annual Meeting of the Association for Computational Linguistics (Volume 1: Long Papers)},
  month     = jul,
  year      = {2023},
  address   = {Toronto, Canada},
  publisher = {Association for Computational Linguistics},
  url       = {https://aclanthology.org/2023.acl-long.634},
  doi       = {10.18653/v1/2023.acl-long.634},
  pages     = {11328--11348}
}

@article{10.1162/coli.a.16,
  author   = {Zhang, Yue and Li, Yafu and Cui, Leyang and Cai, Deng and Liu, Lemao and Fu, Tingchen and Huang, Xinting and Zhao, Enbo and Zhang, Yu and Chen, Yulong and Wang, Longyue and Luu, Ahn Tuan and Bi, Wei and Shi, Freda and Shi, Shuming},
  title    = {Siren’s Song in the AI Ocean: A Survey on Hallucination in Large Language Models},
  journal  = {Computational Linguistics},
  pages    = {1-45},
  year     = {2025},
  month    = {07},
  issn     = {0891-2017},
  doi      = {10.1162/coli.a.16},
  url      = {https://doi.org/10.1162/coli.a.16},
  eprint   = {https://direct.mit.edu/coli/article-pdf/doi/10.1162/coli.a.16/2535477/coli.a.16.pdf}
}

@article{fabbri2020summeval,
  title   = {SummEval: Re-evaluating Summarization Evaluation},
  author  = {Fabbri, Alexander R and Kry{\'s}ci{\'n}ski, Wojciech and McCann, Bryan and Xiong, Caiming and Socher, Richard and Radev, Dragomir},
  journal = {arXiv preprint arXiv:2007.12626},
  year    = {2020}
}

@article{ravi2024lynx,
  title   = {Lynx: An open source hallucination evaluation model},
  author  = {Ravi, Selvan Sunitha and Mielczarek, Bartosz and Kannappan, Anand and Kiela, Douwe and Qian, Rebecca},
  journal = {arXiv preprint arXiv:2407.08488},
  year    = {2024}
}

@article{kim2025medical,
  title   = {Medical hallucinations in foundation models and their impact on healthcare},
  author  = {Kim, Yubin and Jeong, Hyewon and Chen, Shan and Li, Shuyue Stella and Lu, Mingyu and Alhamoud, Kumail and Mun, Jimin and Grau, Cristina and Jung, Minseok and Gameiro, Rodrigo and others},
  journal = {arXiv preprint arXiv:2503.05777},
  year    = {2025}
}

@article{HERRERATAPIAS20251184,
  title    = {Legal Hallucinations and the Adoption of Artificial Intelligence in the Judiciary},
  journal  = {Procedia Computer Science},
  volume   = {257},
  pages    = {1184-1189},
  year     = {2025},
  note     = {The 16th International Conference on Ambient Systems, Networks and Technologies Networks (ANT)/ the 8th International Conference on Emerging Data and Industry 4.0 (EDI40)},
  issn     = {1877-0509},
  doi      = {https://doi.org/10.1016/j.procs.2025.03.158},
  url      = {https://www.sciencedirect.com/science/article/pii/S1877050925008956},
  author   = {Beliña Annery Herrera-Tapias and Diego {Hernández Guzmán}}
}

@article{liu2025reducing,
  title={Reducing hallucinations of large language models via hierarchical semantic piece},
  author={Liu, Yanyi and Yang, Qingwen and Tang, Jiawei and Guo, Tiezheng and Wang, Chen and Li, Pan and Xu, Sai and Gao, Xianlin and Li, Zhi and Liu, Jun and others},
  journal={Complex \& Intelligent Systems},
  volume={11},
  number={5},
  pages={1--19},
  year={2025},
  publisher={Springer}
}

@article{xu2024hallucination,
  title={Hallucination is inevitable: An innate limitation of large language models},
  author={Xu, Ziwei and Jain, Sanjay and Kankanhalli, Mohan},
  journal={arXiv preprint arXiv:2401.11817},
  year={2024}
}

@article{gao2022rarr,
  title={Rarr: Researching and revising what language models say, using language models},
  author={Gao, Luyu and Dai, Zhuyun and Pasupat, Panupong and Chen, Anthony and Chaganty, Arun Tejasvi and Fan, Yicheng and Zhao, Vincent Y and Lao, Ni and Lee, Hongrae and Juan, Da-Cheng and others},
  journal={arXiv preprint arXiv:2210.08726},
  year={2022}
}

@article{laban2022summac,
  title={SummaC: Re-visiting NLI-based models for inconsistency detection in summarization},
  author={Laban, Philippe and Schnabel, Tobias and Bennett, Paul N and Hearst, Marti A},
  journal={Transactions of the Association for Computational Linguistics},
  volume={10},
  pages={163--177},
  year={2022},
  publisher={MIT Press One Rogers Street, Cambridge, MA 02142-1209, USA journals-info~…}
}

@inproceedings{lei2025factcg,
  title={FactCG: Enhancing Fact Checkers with Graph-Based Multi-Hop Data},
  author={Lei, Deren and Li, Yaxi and Li, Siyao and Hu, Mengya and Xu, Rui and Archer, Ken and Wang, Mingyu and Ching, Emily and Deng, Alex},
  booktitle={Proceedings of the 2025 Conference of the Nations of the Americas Chapter of the Association for Computational Linguistics: Human Language Technologies (Volume 1: Long Papers)},
  pages={5002--5020},
  year={2025}
}

@inproceedings{fabbri2022qafacteval,
  title={QAFactEval: Improved QA-Based Factual Consistency Evaluation for Summarization},
  author={Fabbri, Alexander Richard and Wu, Chien-Sheng and Liu, Wenhao and Xiong, Caiming},
  booktitle={Proceedings of the 2022 Conference of the North American Chapter of the Association for Computational Linguistics: Human Language Technologies},
  pages={2587--2601},
  year={2022}
}

@inproceedings{tang2024minicheck,
  title={MiniCheck: Efficient Fact-Checking of LLMs on Grounding Documents},
  author={Tang, Liyan and Laban, Philippe and Durrett, Greg},
  booktitle={Proceedings of the 2024 Conference on Empirical Methods in Natural Language Processing},
  pages={8818--8847},
  year={2024}
}

@article{wei2022chain,
  title={Chain-of-thought prompting elicits reasoning in large language models},
  author={Wei, Jason and Wang, Xuezhi and Schuurmans, Dale and Bosma, Maarten and Xia, Fei and Chi, Ed and Le, Quoc V and Zhou, Denny and others},
  journal={Advances in neural information processing systems},
  volume={35},
  pages={24824--24837},
  year={2022}
}

@article{wang2022self,
  title={Self-consistency improves chain of thought reasoning in language models},
  author={Wang, Xuezhi and Wei, Jason and Schuurmans, Dale and Le, Quoc and Chi, Ed and Narang, Sharan and Chowdhery, Aakanksha and Zhou, Denny},
  journal={arXiv preprint arXiv:2203.11171},
  year={2022}
}

@article{yao2023tree,
  title={Tree of thoughts: Deliberate problem solving with large language models},
  author={Yao, Shunyu and Yu, Dian and Zhao, Jeffrey and Shafran, Izhak and Griffiths, Tom and Cao, Yuan and Narasimhan, Karthik},
  journal={Advances in neural information processing systems},
  volume={36},
  pages={11809--11822},
  year={2023}
}

@article{guo2025deepseek,
  title={Deepseek-r1: Incentivizing reasoning capability in llms via reinforcement learning},
  author={Guo, Daya and Yang, Dejian and Zhang, Haowei and Song, Junxiao and Zhang, Ruoyu and Xu, Runxin and Zhu, Qihao and Ma, Shirong and Wang, Peiyi and Bi, Xiao and others},
  journal={arXiv preprint arXiv:2501.12948},
  year={2025}
}

@article{shao2024deepseekmath,
  title={Deepseekmath: Pushing the limits of mathematical reasoning in open language models},
  author={Shao, Zhihong and Wang, Peiyi and Zhu, Qihao and Xu, Runxin and Song, Junxiao and Bi, Xiao and Zhang, Haowei and Zhang, Mingchuan and Li, YK and Wu, Yang and others},
  journal={arXiv preprint arXiv:2402.03300},
  year={2024}
}

@article{yang2025qwen3,
  title={Qwen3 technical report},
  author={Yang, An and Li, Anfeng and Yang, Baosong and Zhang, Beichen and Hui, Binyuan and Zheng, Bo and Yu, Bowen and Gao, Chang and Huang, Chengen and Lv, Chenxu and others},
  journal={arXiv preprint arXiv:2505.09388},
  year={2025}
}

@article{zhang2025qwen3,
  title={Qwen3 Embedding: Advancing Text Embedding and Reranking Through Foundation Models},
  author={Zhang, Yanzhao and Li, Mingxin and Long, Dingkun and Zhang, Xin and Lin, Huan and Yang, Baosong and Xie, Pengjun and Yang, An and Liu, Dayiheng and Lin, Junyang and others},
  journal={arXiv preprint arXiv:2506.05176},
  year={2025}
}

@article{zhao2025swift,
  title        = {SWIFT: A Scalable Lightweight Infrastructure for Fine-Tuning},
  volume       = {39},
  url          = {https://ojs.aaai.org/index.php/AAAI/article/view/35383},
  doi          = {10.1609/aaai.v39i28.35383},
  abstractnote = {Recent development in Large Language Models (LLMs) and Multi-modal Large Language Models (MLLMs) have achieved superior performance and generalization capabilities, covered extensive areas of traditional tasks. However, existing large model training frameworks support only a limited number of models and techniques, particularly lacking in support for new models, which makes fine-tuning LLMs challenging for most developers. Therefore, we develop SWIFT, a customizable one-stop infrastructure for large models. With support of over 350+ LLMs and 80+ MLLMs, SWIFT stands as the open-source framework that provide the most comprehensive support for fine-tuning large models. In particular, it is the first training framework that provides systematic support for MLLMs. Moreover, SWIFT integrates post-training processes such as inference, evaluation, and quantization, to facilitate fast adoptions of large models in various application scenarios, offering helpful utilities like benchmark comparisons among different training techniques.},
  number       = {28},
  journal      = {Proceedings of the AAAI Conference on Artificial Intelligence},
  author       = {Zhao, Yuze and Huang, Jintao and Hu, Jinghan and Wang, Xingjun and Mao, Yunlin and Zhang, Daoze and Jiang, Zeyinzi and Wu, Zhikai and Ai, Baole and Wang, Ang and Zhou, Wenmeng and Chen, Yingda},
  year         = {2025},
  month        = {Apr.},
  pages        = {29733-29735}
}

@inproceedings{wolf-etal-2020-transformers,
    title = "Transformers: State-of-the-Art Natural Language Processing",
    author = "Thomas Wolf and Lysandre Debut and Victor Sanh and Julien Chaumond and Clement Delangue and Anthony Moi and Pierric Cistac and Tim Rault and Rémi Louf and Morgan Funtowicz and Joe Davison and Sam Shleifer and Patrick von Platen and Clara Ma and Yacine Jernite and Julien Plu and Canwen Xu and Teven Le Scao and Sylvain Gugger and Mariama Drame and Quentin Lhoest and Alexander M. Rush",
    booktitle = "Proceedings of the 2020 Conference on Empirical Methods in Natural Language Processing: System Demonstrations",
    month = oct,
    year = "2020",
    address = "Online",
    publisher = "Association for Computational Linguistics",
    url = "https://www.aclweb.org/anthology/2020.emnlp-demos.6",
    pages = "38--45"
}

@inproceedings{kwon2023efficient,
  title={Efficient Memory Management for Large Language Model Serving with PagedAttention},
  author={Woosuk Kwon and Zhuohan Li and Siyuan Zhuang and Ying Sheng and Lianmin Zheng and Cody Hao Yu and Joseph E. Gonzalez and Hao Zhang and Ion Stoica},
  booktitle={Proceedings of the ACM SIGOPS 29th Symposium on Operating Systems Principles},
  year={2023}
}

@misc {hhem-2.1-open,
    author       = {Forrest Bao and Miaoran Li and Rogger Luo and Ofer Mendelevitch},
    title        = {{HHEM-2.1-Open}},
    year         = 2024,
    url          = { https://huggingface.co/vectara/hallucination_evaluation_model },
    doi          = { 10.57967/hf/3240 },
    publisher    = { Hugging Face }
}

@article{gu2024survey,
  title={A survey on llm-as-a-judge},
  author={Gu, Jiawei and Jiang, Xuhui and Shi, Zhichao and Tan, Hexiang and Zhai, Xuehao and Xu, Chengjin and Li, Wei and Shen, Yinghan and Ma, Shengjie and Liu, Honghao and others},
  journal={arXiv preprint arXiv:2411.15594},
  year={2024}
}

\clearpage
\section{Appendix}

\subsection{All used open-source models}

In our experiments, we used the following open-source models from the Hugging Face hub. The models we used along with their respective hub tags are listed in Table~\ref{tab:used_models}.
Note that Qwen3-32B uses FP8 precision to reduce deployment costs.

\begin{table}[htbp]
    \centering
    \resizebox{\columnwidth}{!}{
        \begin{tabular}{ll}
            \toprule
            \textbf{Model} & \textbf{Hugging Face Hub Tags}                    \\
            \midrule
            Alignscore-L   & yzha/AlignScore                                   \\
            FactCG-DBT     & yaxili96/FactCG-DeBERTa-v3-Large                  \\
            MiniCheck-FT5  & lytang/MiniCheck-Flan-T5-Large                    \\
            MiniCheck-7B   & bespokelabs/Bespoke-MiniCheck-7B                  \\
            Lynx-8B        & PatronusAI/Llama-3-Patronus-Lynx-8B-Instruct-v1.1 \\
            Qwen3-4B       & Qwen/Qwen3-4B                                     \\
            Qwen3-32B      & Qwen/Qwen3-32B-FP8                                \\
            HHEM           & vectara/hallucination\_evaluation\_model          \\
            \bottomrule
        \end{tabular}
    }
    \caption{List of models used in the experiments and their corresponding Hugging Face hub tags.}
    \label{tab:used_models}

\end{table}
\subsection{Hallucination Diagnosis Baseline Construction}

In paper, we constructed two baselines for the hallucination diagnosis pipeline: the "single-prompt" method and the multi-step "pipeline" method.

The prompt used for the single-step method is exact same as the one used in Section~\ref{sec:hdr4b_rl_template}.

The algorithmic flow of the multi-step pipeline is as follows:

\begin{algorithm}[htbp]
    \caption{Hallucination Diagnosis Pipeline Baseline}
    \label{alg:hallu_fix_abstracted}
    \begin{algorithmic}[1]
        \STATE \textbf{Input:} Premise $P$, Hypothesis $H$
        \STATE $is\_hallucination \gets \text{DetectHallucination}(P, H)$
        \IF{not $is\_hallucination$}
        \STATE \textbf{return} $H$, [ ], $\text{False}$
        \ENDIF
        \STATE $S_{hallu} \gets \text{FindHallucinatedSents}(P, H)$
        \STATE $H_{fixed} \gets \text{FixHypothesis}(P, H)$
        \STATE \textbf{return} $H_{fixed}$, $S_{hallu}$, $is\_hallucination$
    \end{algorithmic}
\end{algorithm}

\subsection{Training Template for HDR-4B-RL}
\label{sec:hdr4b_rl_template}

To train HDR-4B-RL and simplify the extraction of results in reward calculation, we used a training prompt based on JSON Object generation, as follows:

\begin{tcolorbox}[title=System Prompt]
    \#\#\# Role \& Objective
    \newline
    \newline
    You are a **Hallucination Diagnosis Expert**. Your expertise lies in the meticulous, evidence-based analysis of AI-generated answers to identify and explain any deviations from factual accuracy or contextual relevance.
    \newline
    Your mission is to analyze a given `Answer` against its source `Context` and the user's `Query`. Based on this analysis, you will produce a comprehensive Diagnosis Report. If the answer is hallucinated, the conclusion should be "Fail" and you should provide a detailed explanation of the hallucinations. If the answer is correct, the conclusion should be "Pass".
    \newline
    \newline
    \#\#\# Output Format
    \newline
    \newline
    You should output a JSON object with the following keys: \\
    - `conclusion`: A string indicating whether the answer is "Pass" or "Fail". \\
    - `diagnosis`: A string providing a detailed explanation of the analysis, including any hallucinations\\
    - `hallucinations`: A list of strings, each representing a hallucinated segment of the answer, if any. If there are no hallucinations, this list should be empty.\\
    - `corrected\_answer`: A string containing the corrected answer if the original answer is hallucinated, or an empty string if the answer is correct.

\end{tcolorbox}

\begin{tcolorbox}[title=User Prompt]
    \#\#\# Input Data

    Context:\\
    \{context\}
    \newline
    \newline
    Query: \{query\} \\
    Answer: \{answer\}

    \#\#\# Begin Execution
\end{tcolorbox}

\subsection{Full Experiment Result for Hallucination Detection Task}

In the paper, we reported the experimental results for the hallucination detection task using Marco-F1 as the metric. To provide a more comprehensive view of the results, we present all metrics here, including Accuracy, Precision, Recall, and F1-score. Bold and underlined values indicate the best and the second-best results in each group, respectively. Note that a model with $^{\dag}$ indicates it was evaluated in non-reasoning mode (w/o thinking).

\begin{table*}[htbp]
    \centering
    \setlength{\tabcolsep}{3.5pt}

    \begin{tabular}{ll cccc  cccc  cccc}
        \toprule
        \multirow{2}{*}{\textbf{Model}} & \multirow{2}{*}{\textbf{Size}} & \multicolumn{4}{c}{\textbf{HaluEval}} & \multicolumn{4}{c}{\textbf{RAGTruth}} & \multicolumn{4}{c}{\textbf{FinanceBench}}                                                                                                                                                          \\

        \cmidrule(lr){3-6} \cmidrule(lr){7-10} \cmidrule(lr){11-14}

                                        &                                & \textbf{P}                            & \textbf{R}                            & \textbf{F1}                               & \textbf{Acc}   & \textbf{P}     & \textbf{R}     & \textbf{F1}    & \textbf{Acc}   & \textbf{P}     & \textbf{R}     & \textbf{F1}    & \textbf{Acc}   \\
        \midrule

        \multicolumn{14}{l}{\textit{Prompt Based}}                                                                                                                                                                                                                                                                                                            \\
        \cmidrule(lr){1-14}
        Qwen3-4B$^{\dag}$               & 4B                             & 77.92                                 & 75.16                                 & 73.64                                     & 74.44          & 63.87          & 57.16          & 58.25          & 81.11          & 59.07          & 58.00          & 56.72          & 58.00          \\
        Qwen3-4B                        & 4B                             & 75.24                                 & 75.16                                 & 75.15                                     & 75.17          & 62.20          & 60.64          & 61.28          & 78.78          & 83.17          & 83.00          & 82.98          & 83.00          \\
        Qwen3-32B$^{\dag}$              & 32B                            & \textbf{83.75}                        & 81.29                                 & 80.92                                     & 81.26          & 60.44          & 56.99          & 57.83          & 79.22          & 73.75          & 70.60          & 69.59          & 70.60          \\
        Qwen3-32B                       & 32B                            & 77.77                                 & 77.65                                 & 77.63                                     & 77.66          & 67.60          & 74.16          & 69.36          & 78.44          & 86.79          & 86.60          & 86.58          & 86.60          \\
        GPT-4.1                         & -                              & 83.25                                 & 82.99                                 & 82.95                                     & 82.98          & 64.37          & 61.52          & 62.58          & 80.22          & 67.51          & 63.20          & 60.79          & 63.20          \\
        o4-mini                         & -                              & 83.49                                 & \textbf{83.27}                        & \textbf{83.23}                            & \textbf{83.26} & \textbf{69.84} & \textbf{78.51} & \textbf{71.89} & 79.56          & \textbf{88.96} & \textbf{88.90} & \textbf{88.90} & \textbf{88.90} \\
        \midrule

        \multicolumn{14}{l}{\textit{Hallucination Detection Models}}                                                                                                                                                                                                                                                                                          \\
        \cmidrule(lr){1-14}
        HHEM                            & 110M                           & 66.69                                 & 66.63                                 & 66.59                                     & 66.62          & \textbf{75.48} & 72.36          & \textbf{73.72} & \textbf{85.56} & 51.76          & 50.60          & 40.85          & 50.60          \\
        Alignscore-Large                & 355M                           & 74.39                                 & 74.17                                 & 74.10                                     & 74.16          & 57.66          & 56.31          & 56.77          & 76.89          & 50.96          & 50.40          & 41.93          & 50.40          \\
        FactCG-DeBERTa                  & 435M                           & 72.83                                 & 68.03                                 & 66.29                                     & 68.08          & 52.26          & 52.69          & 52.28          & 69.33          & 48.97          & 49.50          & 42.03          & 49.50          \\
        MiniCheck-FT5                   & 783M                           & 67.64                                 & 67.09                                 & 66.85                                     & 67.11          & 53.62          & 54.63          & 53.65          & 68.89          & 62.62          & 53.20          & 42.46          & 53.20          \\
        Bespoke-MiniCheck               & 7B                             & 79.96                                 & 79.76                                 & 79.73                                     & 79.77          & 60.27          & 67.32          & 53.08          & 56.33          & 57.21          & 54.70          & 50.38          & 54.70          \\
        Lynx-8B-Instruct                & 8B                             & \textbf{87.40}                        & \textbf{87.00}                        & \textbf{86.96}                            & \textbf{86.99} & 64.71          & 58.88          & 60.25          & 81.11          & 74.30          & 74.30          & 74.30          & 74.30          \\
        HDM-4B-RL$^{\dag}$              & 4B                             & 84.02                                 & 82.99                                 & 82.84                                     & 82.97          & 63.19          & 64.44          & 63.74          & 77.78          & 59.70          & 58.80          & 57.83          & 58.80          \\
        HDM-4B-RL                       & 4B                             & 84.30                                 & 84.00                                 & 83.96                                     & 83.99          & 66.99          & \textbf{74.86} & 68.52          & 76.78          & \textbf{85.14} & \textbf{84.60} & \textbf{84.54} & \textbf{84.60} \\
        \bottomrule
    \end{tabular}
    \caption{Detection Performance on HaluEval, RAGTruth and FinanceBench benchmarks.}
    \label{tab:result_1}
\end{table*}

\begin{table*}[htbp]
    \centering
    \setlength{\tabcolsep}{3.5pt}

    \begin{tabular}{ll cccc cccc cccc}
        \toprule
        \multirow{2}{*}{\textbf{Model}} & \multirow{2}{*}{\textbf{Size}} & \multicolumn{4}{c}{\textbf{DROP}} & \multicolumn{4}{c}{\textbf{CovidQA}} & \multicolumn{4}{c}{\textbf{PubMedQA}}                                                                                                                                                          \\
        \cmidrule(lr){3-6} \cmidrule(lr){7-10} \cmidrule(lr){11-14}
                                        &                                & \textbf{P}                        & \textbf{R}                           & \textbf{F1}                           & \textbf{Acc}   & \textbf{P}     & \textbf{R}     & \textbf{F1}    & \textbf{Acc}   & \textbf{P}     & \textbf{R}     & \textbf{F1}    & \textbf{Acc}   \\
        \midrule

        \multicolumn{14}{l}{\textit{Prompt Based}}                                                                                                                                                                                                                                                                                                   \\
        \cmidrule(lr){1-14}
        Qwen3-4B$^{\dag}$               & 4B                             & 55.70                             & 54.60                                & 52.29                                 & 54.60          & 81.99          & 78.80          & 78.26          & 78.80          & 76.63          & 70.60          & 68.83          & 70.60          \\
        Qwen3-4B                        & 4B                             & 76.23                             & 76.20                                & 76.19                                 & 76.20          & 88.82          & 87.70          & 87.61          & 87.70          & 80.17          & 78.10          & 77.72          & 78.10          \\
        Qwen3-32B$^{\dag}$              & 32B                            & 71.35                             & 70.80                                & 70.61                                 & 70.80          & 87.64          & 84.30          & 83.94          & 84.30          & 86.02          & 83.70          & 83.43          & 83.70          \\
        Qwen3-32B                       & 32B                            & \textbf{80.12}                    & \textbf{80.10}                       & \textbf{80.10}                        & \textbf{80.10} & 93.14          & 93.00          & 92.99          & 93.00          & 87.10          & 87.10          & 87.10          & 87.10          \\
        GPT-4.1                         & -                              & 67.19                             & 67.00                                & 66.91                                 & 67.00          & 91.66          & 90.80          & 90.75          & 90.80          & \textbf{88.62} & \textbf{88.20} & \textbf{88.17} & \textbf{88.20} \\
        o4-mini                         & -                              & 76.39                             & 76.30                                & 76.28                                 & 76.30          & \textbf{94.41} & \textbf{94.20} & \textbf{94.19} & \textbf{94.20} & 86.53          & 86.00          & 85.95          & 86.00          \\
        \midrule

        \multicolumn{14}{l}{\textit{Hallucination Detection Models}}                                                                                                                                                                                                                                                                                 \\
        \cmidrule(lr){1-14}
        HHEM                            & 110M                           & 51.17                             & 50.80                                & 46.55                                 & 50.80          & 69.40          & 61.70          & 57.48          & 61.70          & 60.32          & 58.80          & 57.22          & 58.80          \\
        Alignscore-Large                & 355M                           & 51.01                             & 51.00                                & 50.90                                 & 51.00          & 74.01          & 66.80          & 64.11          & 66.80          & 68.05          & 63.50          & 61.05          & 63.50          \\
        FactCG-DeBERTa                  & 435M                           & 51.93                             & 51.50                                & 48.67                                 & 51.50          & 82.13          & 79.00          & 78.48          & 79.00          & 71.59          & 63.40          & 59.56          & 63.40          \\
        MiniCheck-FT5                   & 783M                           & 51.61                             & 51.40                                & 49.74                                 & 51.40          & 81.08          & 76.40          & 75.48          & 76.40          & 72.92          & 72.00          & 71.72          & 72.00          \\
        Bespoke-MiniCheck               & 7B                             & 66.25                             & 65.30                                & 64.78                                 & 65.30          & 89.54          & 88.20          & 88.10          & 88.20          & 77.86          & 67.90          & 64.75          & 67.90          \\
        Lynx-8B-Instruct                & 8B                             & 64.81                             & 64.80                                & 64.79                                 & 64.80          & \textbf{97.70} & \textbf{97.70} & \textbf{97.70} & \textbf{97.70} & \textbf{88.76} & \textbf{88.40} & \textbf{88.37} & \textbf{88.40} \\
        HDM-4B-RL$^{\dag}$              & 4B                             & 52.89                             & 52.50                                & 50.85                                 & 52.50          & 85.13          & 84.30          & 84.21          & 84.30          & 80.87          & 80.70          & 80.67          & 80.70          \\
        HDM-4B-RL                       & 4B                             & \textbf{78.25}                    & \textbf{75.10}                       & \textbf{74.39}                        & \textbf{75.10} & 92.56          & 92.50          & 92.50          & 92.50          & 81.14          & 78.30          & 77.79          & 78.30          \\
        \bottomrule
    \end{tabular}
    \caption{Detection Performance on DROP, CovidQA, and PubMedQA benchmarks.}
    \label{tab:results_drop_covid_pubmed}
\end{table*}

\begin{table}[htbp]
    \centering
    \setlength{\tabcolsep}{6pt}

    \resizebox{\linewidth}{!}{
        \begin{tabular}{ll cccc}
            \toprule
            \multirow{2}{*}{\textbf{Model}} & \multirow{2}{*}{\textbf{Size}} & \multicolumn{4}{c}{\textbf{Summeval}}                                                    \\
            \cmidrule(lr){3-6}
                                            &                                & \textbf{P}                            & \textbf{R}     & \textbf{F1}    & \textbf{Acc}   \\
            \midrule

            \multicolumn{6}{l}{\textit{Prompt Based}}                                                                                                                   \\
            \cmidrule(lr){1-6}
            Qwen3-4B$^{\dag}$               & 4B                             & 81.35                                 & 63.16          & 66.43          & 85.25          \\
            Qwen3-4B                        & 4B                             & 79.38                                 & 74.11          & 76.26          & 87.00          \\
            Qwen3-32B$^{\dag}$              & 32B                            & 81.70                                 & 68.92          & 72.61          & 86.69          \\
            Qwen3-32B                       & 32B                            & 77.98                                 & \textbf{78.35} & \textbf{78.16} & 86.81          \\
            GPT-4.1                         & -                              & \textit{82.30}                        & 71.92          & 75.41          & \textit{87.50} \\
            o4-mini                         & -                              & \textbf{83.29}                        & 74.69          & 77.88          & \textbf{88.38} \\
            \midrule

            \multicolumn{6}{l}{\textit{Hallucination Detection Models}}                                                                                                 \\
            \cmidrule(lr){1-6}
            HHEM                            & 110M                           & \textit{82.81}                        & 59.03          & 61.00          & 84.31          \\
            Alignscore-Large                & 355M                           & \textbf{84.53}                        & 62.35          & 65.60          & 85.44          \\
            FactCG-DeBERTa                  & 435M                           & \textit{82.81}                        & 59.03          & 61.00          & 84.31          \\
            MiniCheck-FT5                   & 783M                           & 80.09                                 & 73.70          & 76.20          & \textbf{87.19} \\
            Bespoke-MiniCheck               & 7B                             & 78.67                                 & \textbf{78.45} & \textbf{78.56} & \textbf{87.19} \\
            Lynx-8B-Instruct                & 8B                             & 66.87                                 & 71.71          & 68.39          & 78.12          \\
            HDM-4B-RL$^{\dag}$              & 4B                             & 68.91                                 & 75.48          & 70.75          & 79.13          \\
            HDM-4B-RL                       & 4B                             & 74.38                                 & 77.95          & 75.88          & 84.44          \\
            \bottomrule
        \end{tabular}
    }
    \caption{Performance results on the Summeval benchmark.}
    \label{tab:results_summeval}
\end{table}

\begin{table}[htbp]
    \centering

    \begin{tabular}{lccccc}
        \toprule
        \textbf{Model}     & \textbf{Size} & \textbf{P}        & \textbf{R}        & \textbf{F1}       & \textbf{ACC}      \\
        \midrule

        \multicolumn{6}{l}{\textit{Prompt Based}}                                                                          \\
        \cmidrule(lr){1-6}
        Qwen3-4B$^{\dag}$  & 4B            & 70.93             & 65.35             & 64.92             & 71.83             \\
        Qwen3-4B           & 4B            & 77.89             & 76.42             & 76.74             & 80.85             \\
        Qwen3-32B$^{\dag}$ & 32B           & 77.81             & 73.80             & 72.58             & 79.51             \\
        Qwen3-32B          & 32B           & \underline{81.50} & \underline{82.42} & \underline{81.70} & \underline{84.24} \\
        GPT-4.1            & -             & 77.84             & 75.09             & 75.37             & 79.99             \\
        o4-mini            & -             & \textbf{83.27}    & \textbf{83.12}    & \textbf{82.62}    & \textbf{85.23}    \\
        \midrule

        \multicolumn{6}{l}{\textit{Hallucination Detection Models}}                                                        \\
        \cmidrule(lr){1-6}
        HHEM               & 110M          & 65.38             & 59.99             & 57.63             & 65.48             \\
        Alignscore-L       & 355M          & 65.80             & 60.65             & 59.21             & 66.88             \\
        FactCG-DBT         & 435M          & 66.07             & 60.45             & 58.33             & 66.45             \\
        MiniCheck-FT5      & 783M          & 67.08             & 64.06             & 62.30             & 68.03             \\
        MiniCheck-7B       & 7B            & 72.82             & 71.66             & 68.48             & 71.34             \\
        Lynx-8B            & 8B            & \underline{77.79} & \underline{77.54} & \underline{77.25} & \underline{81.63} \\
        HDM-4B-RL$^{\dag}$ & 4B            & 70.67             & 71.32             & 70.13             & 73.74             \\
        HDM-4B-RL          & 4B            & \textbf{80.39}    & \textbf{81.04}    & \textbf{79.65}    & \textbf{82.24}    \\
        \bottomrule
    \end{tabular}
    \caption{Average results on all detection benchmark sets.}
    \label{tab:result_all}
\end{table}

\end{document}